\documentclass[10pt,a4paper,conference]{IEEEtran}
\usepackage[utf8]{inputenc}

\usepackage{subfiles}
\usepackage{algorithm}
\usepackage{amsfonts}
\usepackage{algpseudocode}
\usepackage{graphicx}
\usepackage{caption}
\usepackage{subcaption}
\usepackage{amsmath}
\usepackage{mathtools}
\usepackage{algpseudocode}
\usepackage{authblk}
\usepackage{multirow}

\title{Robust Skeletonization for Plant Root Structure Reconstruction from MRI}
\author[1]{Jannis Horn}
\author[1]{Yi Zhao}
\author[1]{Nils Wandel}
\author[2]{Magdalena Landl}
\author[2]{Andrea Schnepf}
\author[1]{Sven Behnke}
\affil[1]{Autonomous Intelligence Systems Group, University of Bonn}
\affil[2]{Forschungszentrum Jülich GmbH, Agrosphere (IBG-3)}
                    
\date{February 2020}

\begin{document}

\maketitle

\begin{abstract}
% I would keep it a little more general, to make it interesting for a broader audience...:
%Skeletonization of noisy input data is deemed to be difficult as disconnectivities can render traditional skeletonization algorithms useless (are there other papers, that tried to close gaps in skeletonization algos or are we the first?).
%NMRooting also allows gap closing 

%Usually before skeletonization morphologic closing operations are applied. This cannot work for our specific example because even small noise elements can add unwanted structures and larger scale closing operations merge roots.

%We propose a new robust skeletonization algorithm to overcome gaps...
% and apply this algorithm to MRI data of plant roots.
% TODO: cite paper, which applied the algorithm to lung vessels
% -> here, we can continue to describe MRIs of plant roots

%To further the understanding of plant root interaction with soil it is necessary to have knowledge of the root structure. MRI can be used to depict plant roots in soil. These scans can suffer from high noise levels and missing connections. To address this an extraction algorithm is proposed. The extractor uses a two-stage approach to first limit connections to non-root noise and to bridge disconnected root segments. In the second stage curve skeletonization applied on the first stage output generates a tree graph containing position and radius information. TODO Results

Structural reconstruction of plant roots from MRI is challenging, because of low resolution and low signal-to-noise ratio of the 3D measurements which may lead to disconnectivities and wrongly connected roots. 
We propose a two-stage approach for this task.
The first stage is based on semantic root vs. soil segmentation and finds lowest-cost paths from any root voxel to the shoot.
The second stage takes the largest fully connected component generated in the first stage and uses 3D skeletonization to extract a graph structure. We evaluate our method on 22 MRI scans and compare to human expert reconstructions.

\end{abstract}

\section{Introduction}

% I would put motivation / problem description first
% e.g.: why do we want to extract roots in MRI data? why do we want to use skeletonization algorithms? what are the problems with gaps / morphological closing? (e.g. large structural elements would destroy morphology...)
% then: sth like: "we propose a novel robust curve skeletonization algorithm, that can overcome gaps of imperfect input data." (I wouldn't go into details, here)

%%% Deleted by Andrea 
% root performance plays a crucial role in understanding plant behaviour given for example different soil types. Therefore plant roots and their interaction in soil should be observed. Magnet Resonance Imaging (MRI), Computer Tomography (CT) \cite{metzner2015direct} or Neutron Radiography (NR) \cite{moradi2009neutron} can be used to depict plant roots in opaque soil in-situ.
%%%
%%% Added by Andrea 
Plant root system architecture is important for plant performance, particularly under challenging environmental conditions such as droughts. 3D volumetric imaging methods such as Magnet Resonance Imaging (MRI)~\cite{stingaciu2013in}, Computer Tomography (CT)~\cite{metzner2015direct} or Neutron Radiography (NR)~\cite{moradi2009neutron} enable in-situ observations of root system development in opaque soil. 
%%%

\begin{figure*}[!t]
    \centering
    \subfloat[Thresholded root MRI]{\includegraphics[width=.24\textwidth]{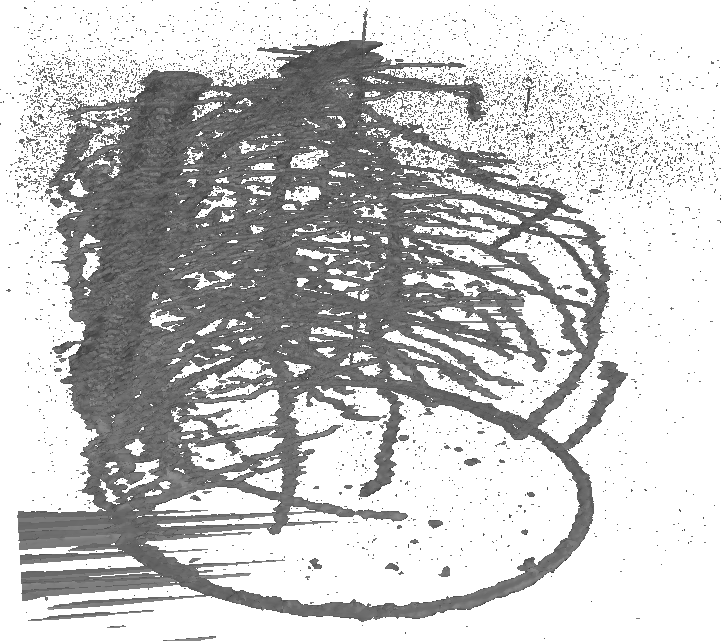}\label{fig:intr_raw}}
    \hfill
    \subfloat[U-Net segmentation]{\includegraphics[width=.24\textwidth]{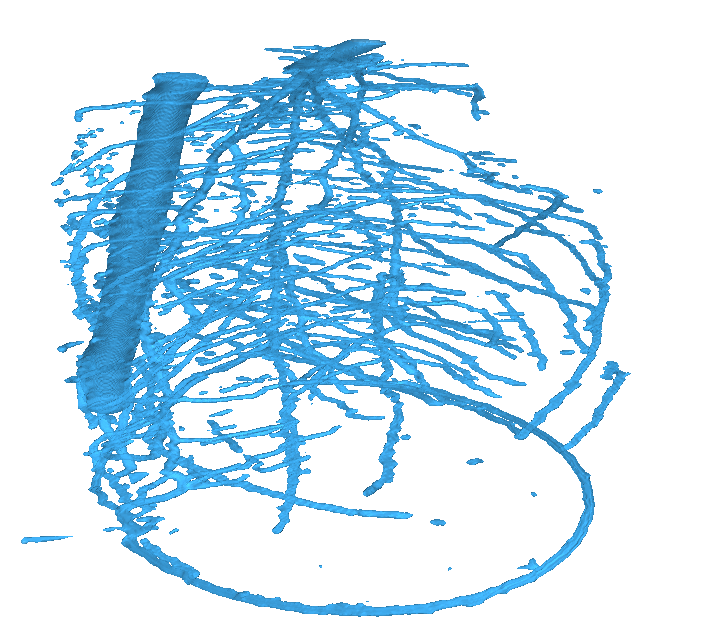}\label{fig:intr_seg}}
    \hfill
    \subfloat[Extracted LCC]{\includegraphics[width=.24\textwidth]{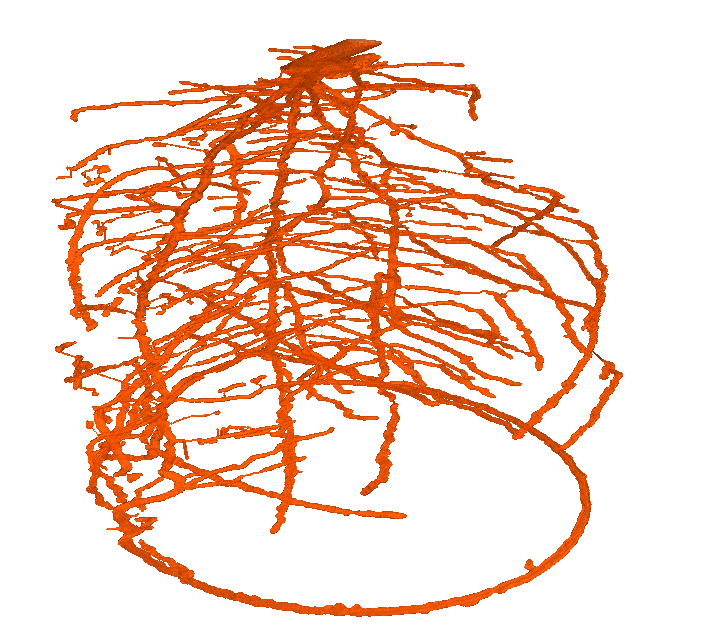}\label{fig:intr_lcc}}
    \hfill
    \subfloat[Root structure graph]{\includegraphics[width=.24\textwidth]{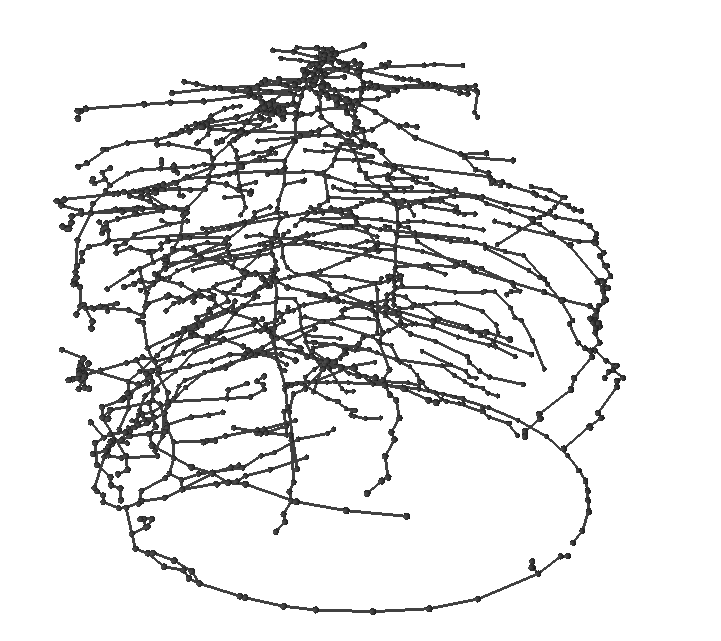}\label{fig:intr_extr}}
		\vspace*{1ex}
    \caption{Overview of root structure extraction. MRI (a) is segmented using a 3D U-Net~\cite{zhao20203d} (b). The largest component connected to the shoot is identified (test-tube removed manually) (c). Gaps are closed and the root structure graph is extracted (d).}
    \label{fig:intr}
\end{figure*}

In this work, MRI images are used which can suffer from low resolution, compared to the diameter of thin roots, and low signal-to-noise ratio (SNR), caused e.g. by ferromagnetic particles in the soil. Fig.~\ref{fig:intr_raw} shows a raw MRI image after thresholding. In Fig.~\ref{fig:intr_seg}, the image has been segmented using a 3D U-Net~\cite{zhao20203d} reducing the noise considerably, but some patches of noise are still present and some roots have gaps.

For plant root system analysis, the structure of a root has to be extracted, i.e. the 3D image has to be transformed into a tree graph structure. Currently, this is done by human experts in a 3D work bench~\cite{stingaciu2013in}, which is a time-consuming process that limits plant root analysis to few samples. Automating root structure extraction allows for fast and reproducible processing of larger MRI measurement sets.

For finding the medial axis of roots, skeletonization algorithms~\cite{jin2016robust} can be used, but gaps must be closed before they can be applied.
For closing gaps in structures, often morphological operations are used. These are sufficient to fill holes and close smaller gaps, but when roots touch each other, closing can lead to wrong connections of root parts. 

\begin{figure}
	\centering
	\includegraphics[width=.48\textwidth]{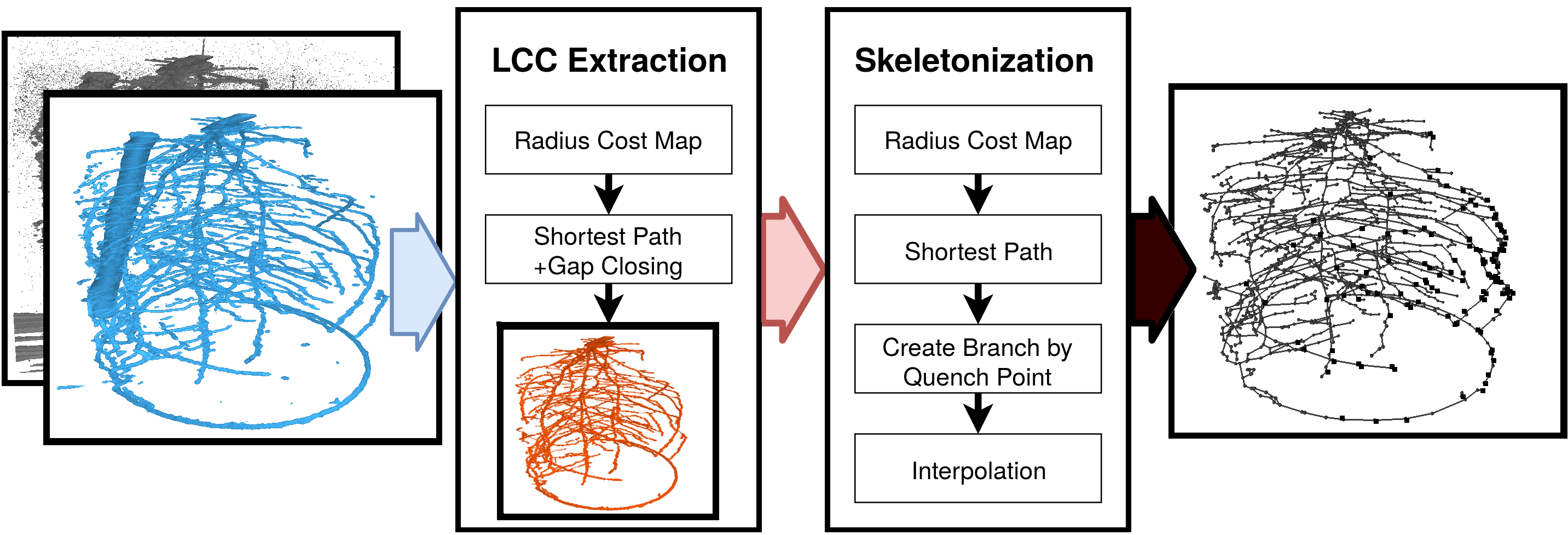}
	\caption{Given a (segmented) plant root MRI scan, first the largest root connected component (LCC) is extracted, followed by skeletonization of the LCC to create a root structure graph.}
\end{figure}

%TODO improve further
In this work, we present a novel robust curve skeletonization algorithm for 3D images. We take a two-stage approach: Stage~1 extracts the largest connected component (LCC), see Fig.~\ref{fig:intr_lcc}, starting from a given set, i.e. the shoot. To this end, a modified Dijkstra shortest path algorithm~\cite{dijkstra1959note} is employed. The shortest path algorithm is modified to connect disconnected areas by finding a single smallest gap. In Stage~2, a curve skeletonization algorithm is applied to the LCC. The resulting algorithm works on imperfect data containing clusters of noise, disconnectivities and intricate object structures. Fig.~\ref{fig:intr_extr} shows the resulting root structure graph.

The presented curve skeletonization algorithm has low computational demands. It is capable of reconstructing the root structure from large 3D images (396$\times$512$\times$512) on modest hardware in short time. We compare the obtained reconstructions to human expert reconstructions.

\section{Related Work}

% here comes some previous work in the field of:
% - plant root imaging
% - root structure extraction
% - skeletonization

%... mention Dakai Jin et Al.: "A Robust and Efficient Curve Skeletonization Algorithm for Tree-Like Objects Using Minimum Cost Paths" (https://www.ncbi.nlm.nih.gov/pmc/articles/PMC4860741/)
%comment Magdalena: I think the terms 'shoot' and 'leaf' should be replaced 

An early root structure extraction algorithm has been developed by Schulz et al. \cite{schulz2013plant}. They used a four-step approach. First, tubular structures at multiple scales are extracted followed by automated plant shoot extraction. Then the Dijkstra shortest path method~\cite{dijkstra1959note} is used to determine connectivity to a set of possible leaf candidates, followed by graph pruning. Pruning is done by deleting branches crossing gaps which are determined as too long and by deleting multiple parallel branches corresponding to the same thicker root.
%%% by Andrea 
The automated root tracking algorithm of Leitner et al.~\cite{Leitner2014} used a mechanistic root growth model to track roots
in a graph representing 2D root systems obtained from NR images. Manual corrections were possible; in that case the Dijkstra algorithm was used to find the shortest path between two manually selected points. 
%%% 
Building on Schulz et al., NMRooting \cite{van2016quantitative} was developed. This tool is implemented using the Python programming language \cite{van2003python} and Mayavi for visualization \cite{ramachandran2011mayavi}. NMRooting extracts a root skeleton by thresholding the input image by a noise-dependent threshold, followed by dilation to bridge smaller gaps. All other voxels above a given threshold are connected using the Dijkstra shortest path algorithm.

These methods are designed to work with high-quality 3D images. To improve MRI scans with insufficient resolution and SNR, deep learning-based methods using a 2D RefineNet have been employed~\cite{uzman2019learning}. Zhao et al.~\cite{zhao20203d} developed a 3D U-Net incorporating further input channels and loss modifications to increase resolution and SNR. The resulting super-resolution segmentation still contains noise and disconnectivities. Hence, a root structure extraction method capable of dealing with noisy data is needed.

Graph shortest path-based extraction methods are widely used in medical imaging as well, e.g. \cite{schafer2007planning} or \cite{zalesky2008dt}. Lung airway extraction is similar to the root structure extraction problem in computing a 3D tree graph from a given 3D image. Jin et al. \cite{jin2016robust} use curve skeletonization based on radius quench points. To reduce the number of false subbranches, the extracted skeletal branches are dilated based on a radius estimate. Quench points inside this dilated branch are ignored. This ensures that for each airway branch only one graph branch is extracted. We follow this basic approach in our work.

For structural analysis of plant roots, voxel-perfect localization of the graph nodes is not necessary. To reduce the number of nodes along elongated roots without deviating much from the original structure, the Douglas-Peucker algorithm \cite{douglas1973algorithms} has proven to work efficiently.

\section{Root Graph Extraction Pipeline}

\begin{figure*}[!t]
	\centering
	\subfloat[Max $I_{seg}$ along depth axis]{\includegraphics[width=.24\textwidth]{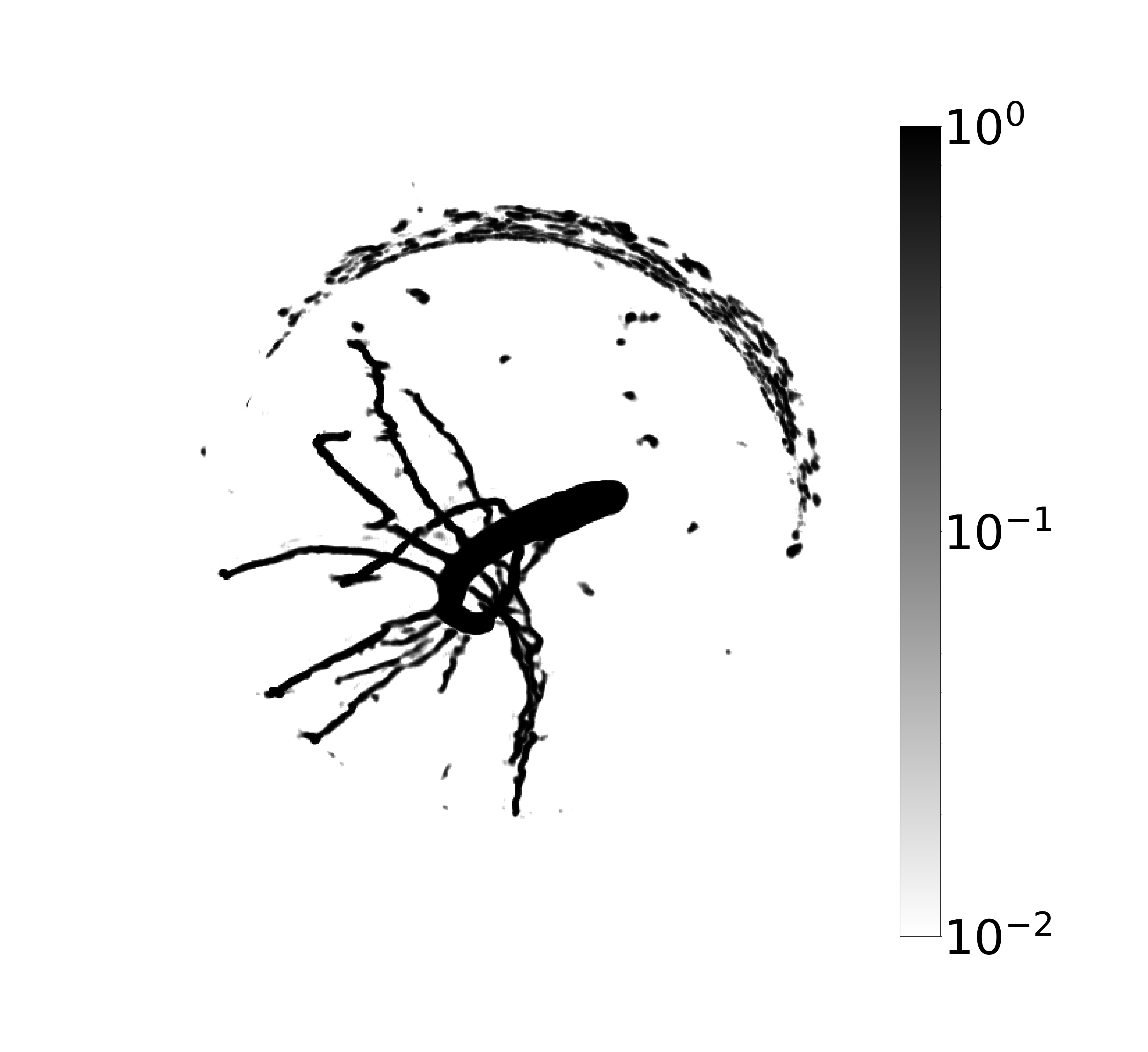}\vspace*{-2ex}\label{fig:run_inp}}
	\hfill
	\subfloat[Min $C_{I_{seg}}$ along depth axis]{\includegraphics[width=.24\textwidth]{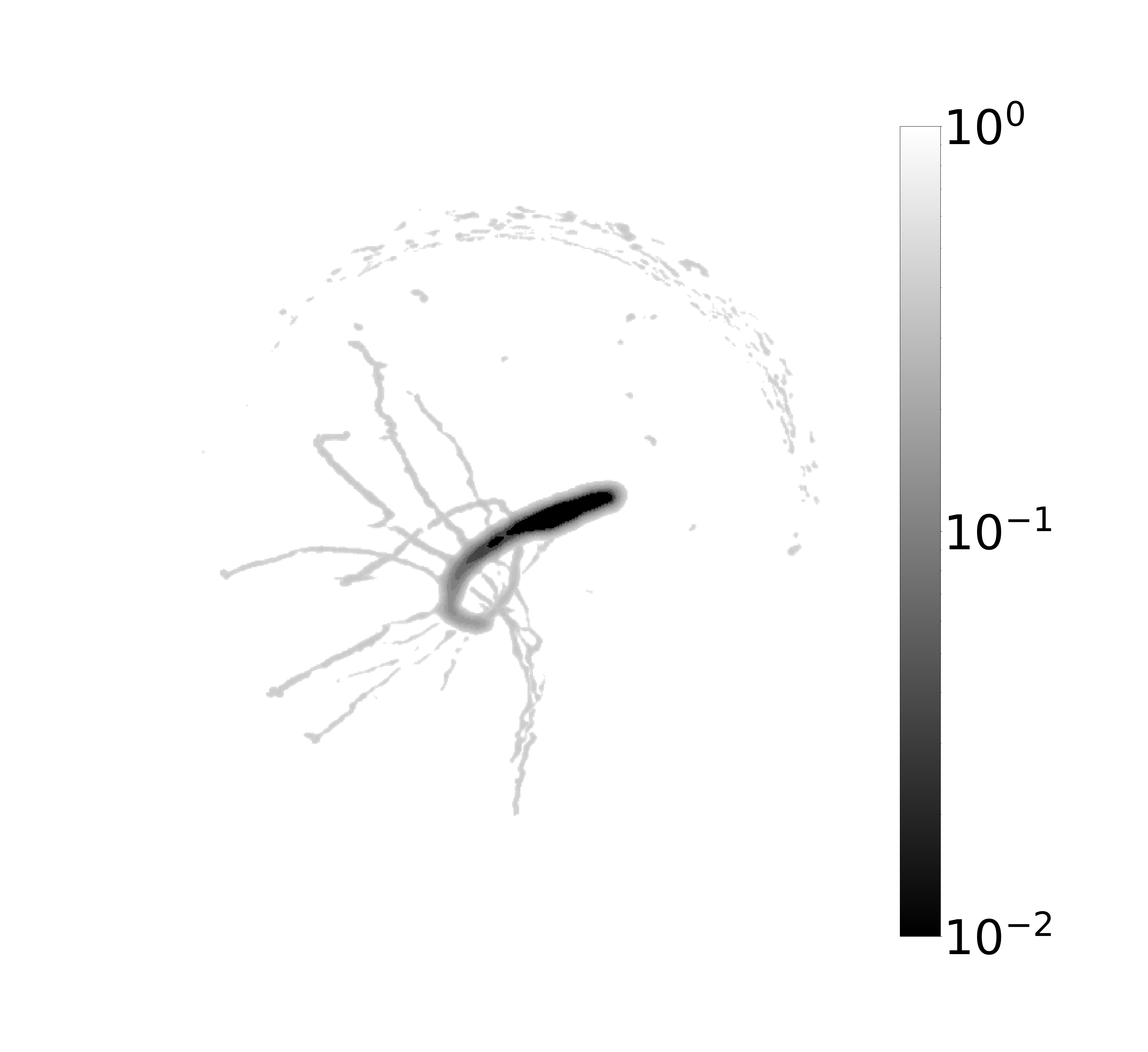}\vspace*{-2ex}\label{fig:run_cost}}
	\hfill
	\subfloat[Min $C_{I_{gap}}$ along depth axis]{\includegraphics[width=.24\textwidth]{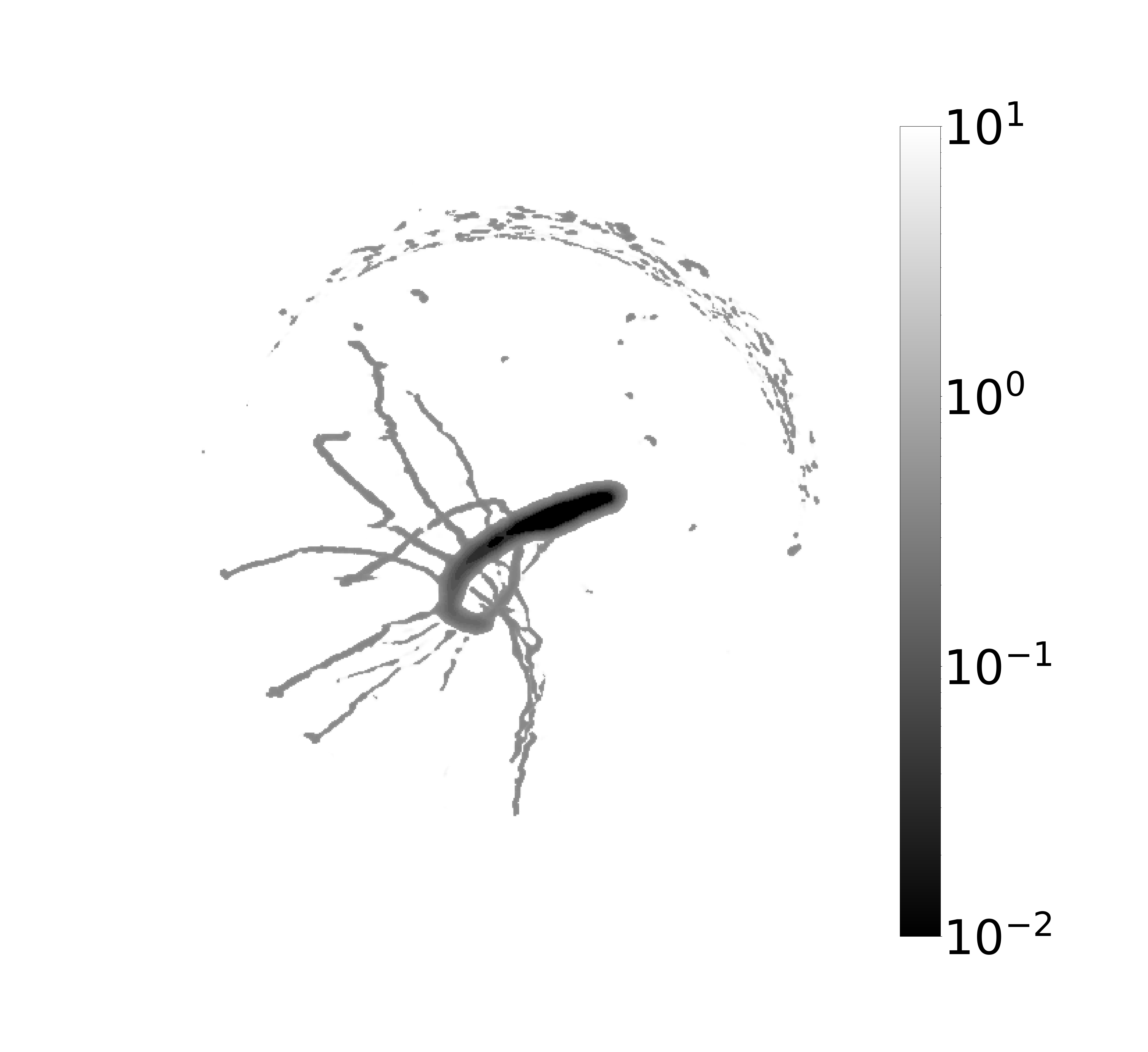}\vspace*{-2ex}\label{fig:run_gap}}
	\hfill
	\subfloat[Max $I_{lcc}$ along depth axis]{\includegraphics[width=.24\textwidth]{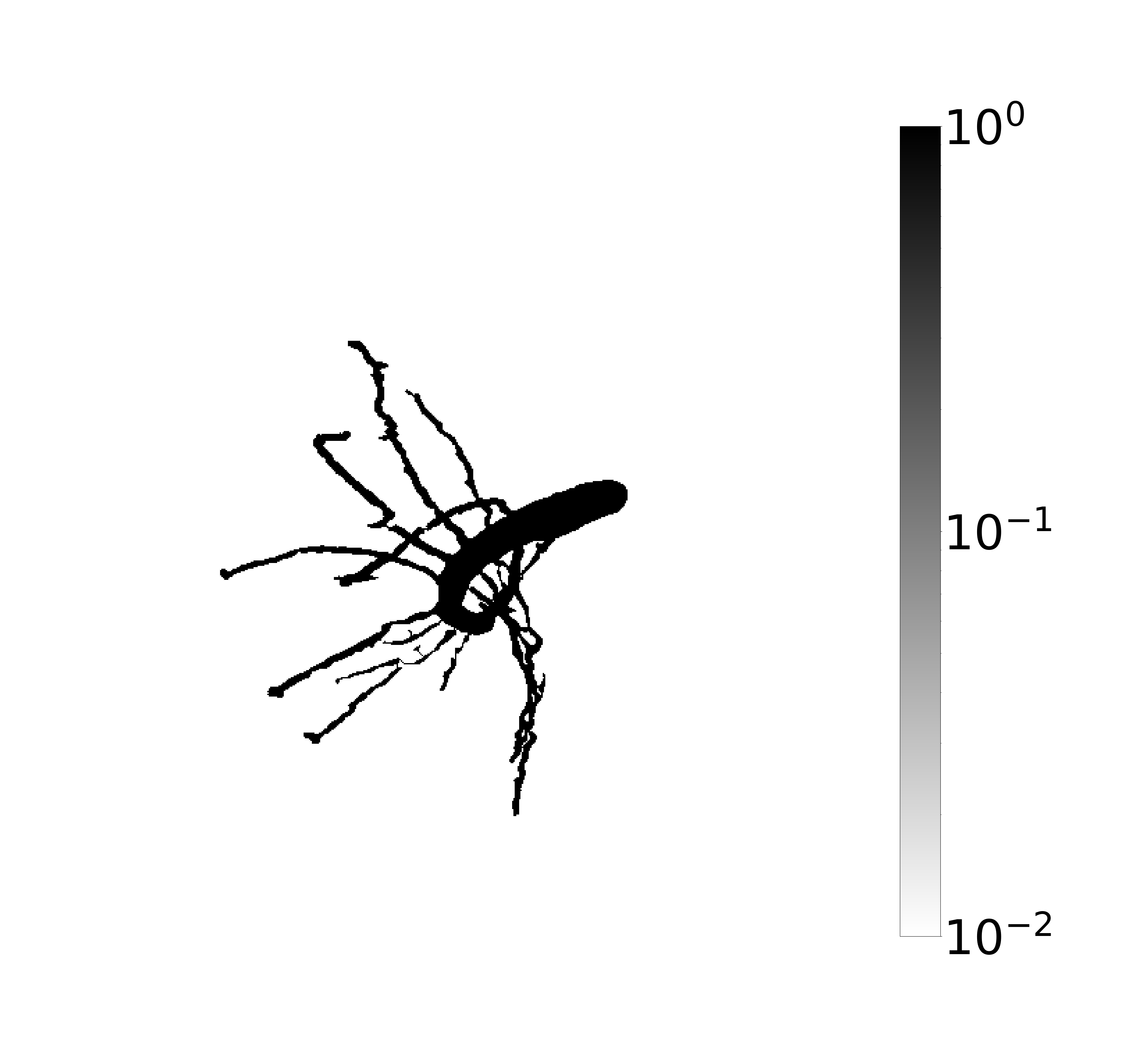}\label{fig:run_lcc}\vspace*{-2ex}}
	\hfill
	\caption{LCC extraction. Given input volume in intensity (a), voxel-cost (b) is computed reducing smaller noise clusters. Cost is masked using $I_{seg}$ (c) to generate $C_{I_{gap}}$ enhancing no-gap to gap contrast. Based on this, $I_{lcc}$ is extracted (d).}
	\label{fig:run_extr}
	\vspace*{1ex}
\end{figure*}

Plant root MRI can suffer from low signal-to-noise ratio and loss of root information. The resulting scans contain non-root signals and disconnected root signals. The first issue can be reduced by employing image segmentation and super-resolution methods (e.g. \cite{uzman2019learning} and \cite{zhao20203d}). Resulting segmented MRI scans $I_{seg}$ still contain a number of noise-signal clusters and disconnected root structures. Current root image segmentation does not take connectivity towards the plant shoot into account. Using this metric further reduces noise clusters in the segmented input by excluding voxels that don't have a low-cost root path towards the shoot.

The presented algorithm employs this connectivity measure to further improve segmentation given a plant root MRI scan $I$ and its initial segmentation $I_{seg}$. In the first step, a largest-connected-component (LCC) $I_{lcc}$ is extracted using $I_{seg}$. To address disconnected root components, the LCC extraction allows for small gaps in the root signal to be bridged. If high-quality input data is used, this step can be omitted.

Noise and inaccuracies in $I$ and subsequently $I_{seg}$ prohibit the use of raw input signal to find the centerline of tubular structures. Higher resolution and noise increase the number and length of subbranches when using an approach similar to \cite{schulz2013plant}. By employing a robust 3D curve skeletonization algorithm based on \cite{jin2016robust}, the number of artificial subbranches and the subsequent amount of necessary computational resources is reduced.

In the following, $G$ is used for graphs. Other upper-case variables are volumes of the same size as $I_{seg}$. Lower-case variables $p$ and $q$ are used instead of full 3D positions $(x,y,z)$.

\section{Largest Connected Component Extraction}
\label{sec:lcc}

Given a segmented input volume $I_{seg} \in [0,1]^{x\cdot y\cdot z}$, a start point $p_{0} \in I_{seg}$ is found either manually or by fitting circle masks in the upper part of the scan. Also given are associated minimum intensity $\gamma_{I_{seg}} \in [0,1]$ and maximum path cost $\omega_{I_{seg}} \in \mathbb{R}$. If gap closing is used, a maximum gap length $l_{I_{seg}} \in \mathbb{N}$ is provided. 

	\subsection{Radius Cost Map}
	\label{sec:rad_cost}
	At each 3D position $p \in I_{seg}$, the radius is estimated by fitting growing spheres centered around $p$ as follows: Let $\phi_r$ be the set of discrete positions inside a sphere of radius r and $\xi_{I_{seg},r}(p) \subset I_{seg}$ be the set of voxels inside the same sphere centered around $p$ and $\xi_{I_{seg},r}^+(p)$ all $p' \in \xi_{I_{seg},r}(p)$ for which volume $I_{seg}(p') \geq \gamma_{I_{seg}}$ then:
	\begin{align}
		R_{I_{seg}}(p) &= \underset{r \in \mathbb{N}}{\text{argmax}}\left(\frac{|\xi_{I_{seg},r'}^+(p)|}{|\phi_{r'}|} \geq 0.75, \forall r' \leq r\right) \label{for:rad_est}, \\
		C_{inv}(p) &= I_{seg}(p) + w_{rad}\frac{R_{I_{seg}}(p)}{\text{max}(R_{I_{seg}})}, \nonumber \\
		C_{I_{seg}}(p) &= 1-\frac{C_{inv}(p)}{\text{max}(C_{inv})} + \epsilon. \nonumber
	\end{align}
	$w_{rad} \in [0,1]$ is the weighting of radius vs. input intensity, $w_{rad} =0.5$ works well in most cases. $\epsilon \in \mathbb{R}_{>0} \ll 1$ is a small constant to avoid zero-cost voxels. The resulting cost map takes both local radius information and intensity into account, see Fig.~\ref{fig:run_cost}.

	\subsection{Shortest Path with Gap Closing}
	The Dijkstra Shortest Path algorithm is used to find the shortest path $\tau(p)$ for each $p \in I_{seg}$ to $p_{0}$, producing the path cost volume $C_{\tau}$. In each iteration, the 26-neighborhood of the visited voxel is expanded. As only voxels with path cost  $<\omega_{I_{seg}}$ are included in the resulting LCC, exploration is stopped once voxels of cost $\ge\omega_{I_{seg}}$ are reached. If gap closing is used, this is ignored for potential gaps of maximum $l_{I_{seg}}$ length.
	
	Using $C_{I_{seg}}$ directly can lead to disconnected areas in $I_{seg}$ being connected along multiple parallel paths through high-cost space. These paths would be part of the LCC creating unwanted structures. This is due to the Dijkstra algorithm only minimizing path cost per voxel, but not the sum of all traversals. 
	
	%\begin{figure}[t!]
	%	\centering
	%	\begin{subfigure}[b]{0.20\textwidth}
	%		\centering
	%		\includegraphics[width=\textwidth]{imgs/fig_seg/no_gap_c.png}
	%		\caption{Black: LCC without using gap closing}
	%		\label{fig:lcc_nogap}
	%	\end{subfigure}
	%	\begin{subfigure}[b]{0.20\textwidth}
	%		\centering
	%		\includegraphics[width=\textwidth]{imgs/fig_seg/gap_c.png}
	%		\caption{Blue: LCC with using gap closing}
	%		\label{fig:lcc_gap}
	%	\end{subfigure}
	%	\caption{LCC extraction without gap closing (a) creates multiple connections outside the input volume. Using gap closing (b) reduces these connections.}
	%\end{figure}
	
	To address this, $C_{I_{seg}}$ is updated:
	\begin{align}
		C_{gap}(p) = 
			\begin{cases}
				C_{I_{seg}}(p) \cdot 10 & \text{, if } I_{seg}(p) < \gamma_{I_{seg}} \\
				C_{I_{seg}}(p) & \text{, otherwise}
			\end{cases}. \label{for:gap}
	\end{align}
	As can be seen in Fig.~\ref{fig:run_gap} this modification increases the contrast between gap and no-gap information. This increases the stability of the gap closing modification for the Dijkstra shortest path.
	
	Disconnected root parts should be connected along a single path per root instead of per voxel. To this end, all positions $p$ where $C_{gap}(p) > \frac{\text{max}(C_{gap})}{2}$ are considered as potential gaps. These are all the areas penalized in Eq.~\ref{for:gap}. The penalty also induces further exploration of no-gap positions before exploration of gap areas. 
	
	\label{sec:gap}

	\begin{figure}[!t]
		\centering
		%\includegraphics[width=.49\textwidth]{imgs/fig_run_cskel/skel_c.png}
		%\subfloat[Extracted $I_{lcc}$]{\includegraphics[width=.24\textwidth]{imgs/fig_run_cskel/lcc_vol.png}\label{fig:run_slcc}}
		%\hfill
		\subfloat[Max normalized $R_{lcc}$]{\includegraphics[width=.24\textwidth]{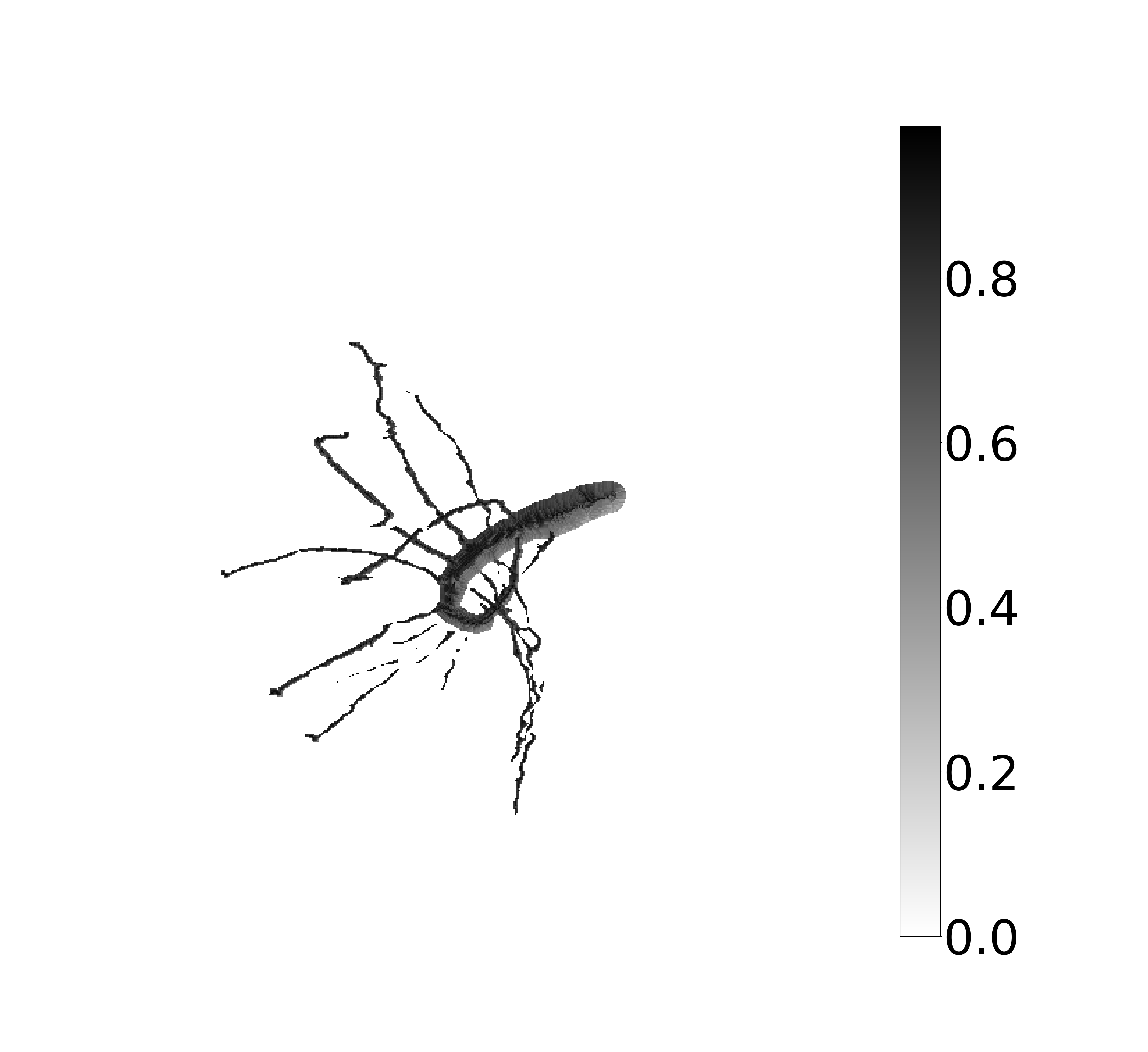}\vspace*{-2ex}\label{fig:run_rad}}
		\hfill
		\subfloat[Max $C_{rel}$ along depth axis]{\includegraphics[width=.24\textwidth]{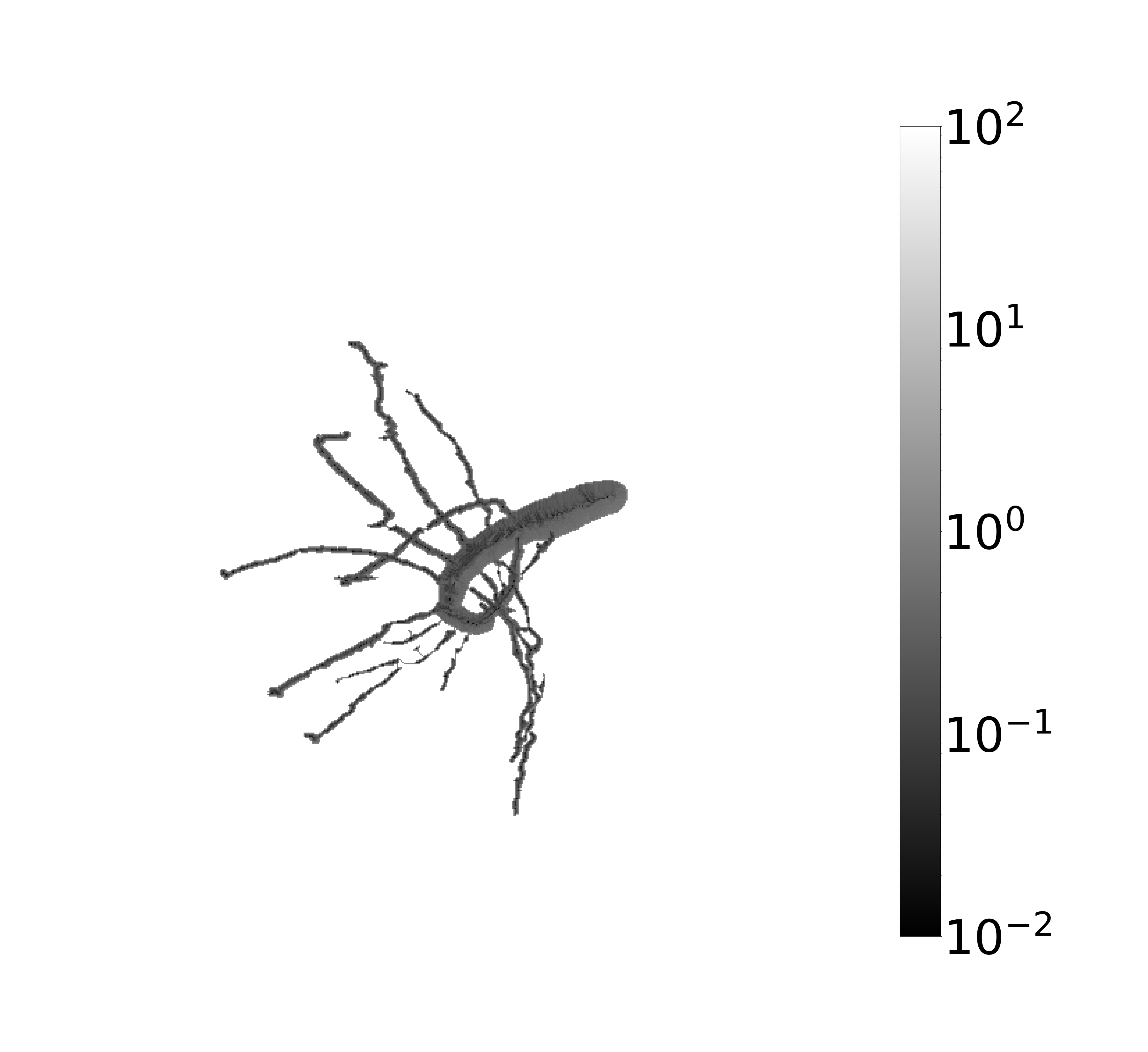}\vspace*{-2ex}\label{fig:run_scost}}
		%\hfill
		%\subfloat[$G$ after full extraction ]{\includegraphics[width=.24\textwidth]{imgs/fig_run_cskel/skel.png}\label{fig:run_graph}}
		\caption{Using the LCC from Fig.~\ref{fig:run_extr}, the radius cost map $R_{lcc}$ (a) is computed. This is inverted and used to create $C_{rel}$ (b).}
		\label{fig:run_cskel}
		%\vspace*{1ex}
	\end{figure}

	\begin{figure*}[!t]
		\centering
		\subfloat[The marked area contains a found \hspace*{3.7ex}quench point (green) for extraction]{\includegraphics[width=.29\textwidth]{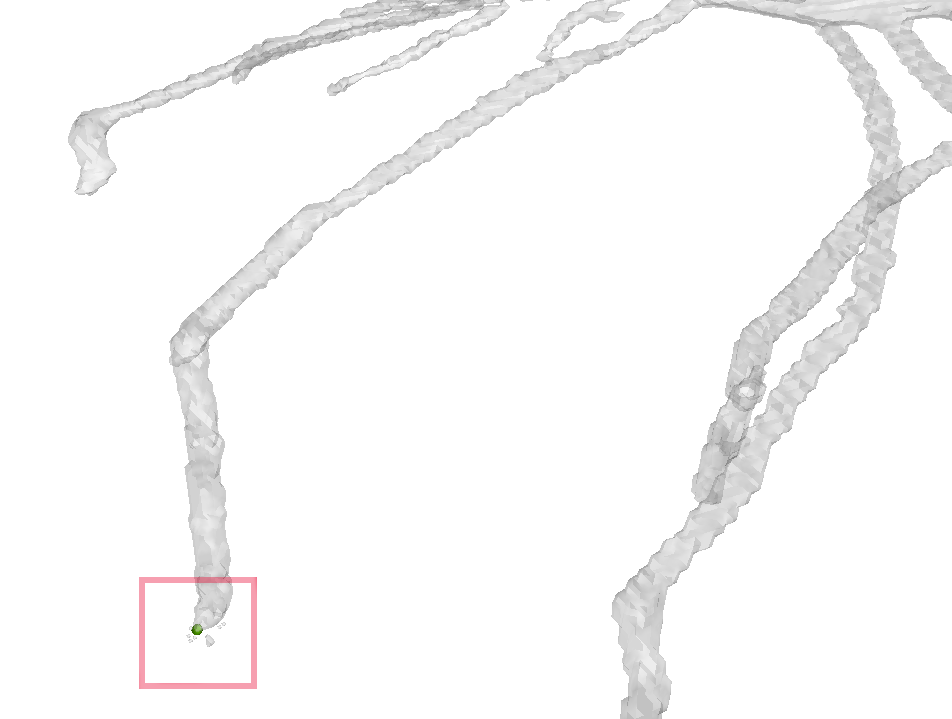}\label{fig:skel_clean}}
		\hfill
		\subfloat[Shortest Path quench point to start \hspace*{3.7ex}point (green, interpolated) ]{\includegraphics[width=.29\textwidth]{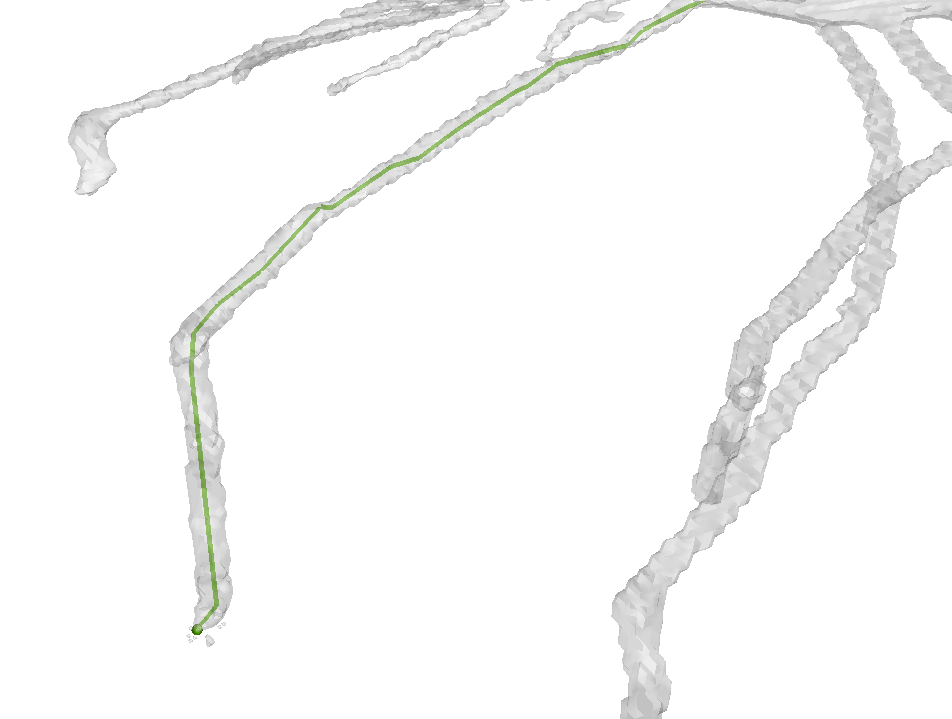}\label{fig:skel_extr}}
		\hfill
		\subfloat[Filled area around extracted path.]{\includegraphics[width=.29\textwidth]{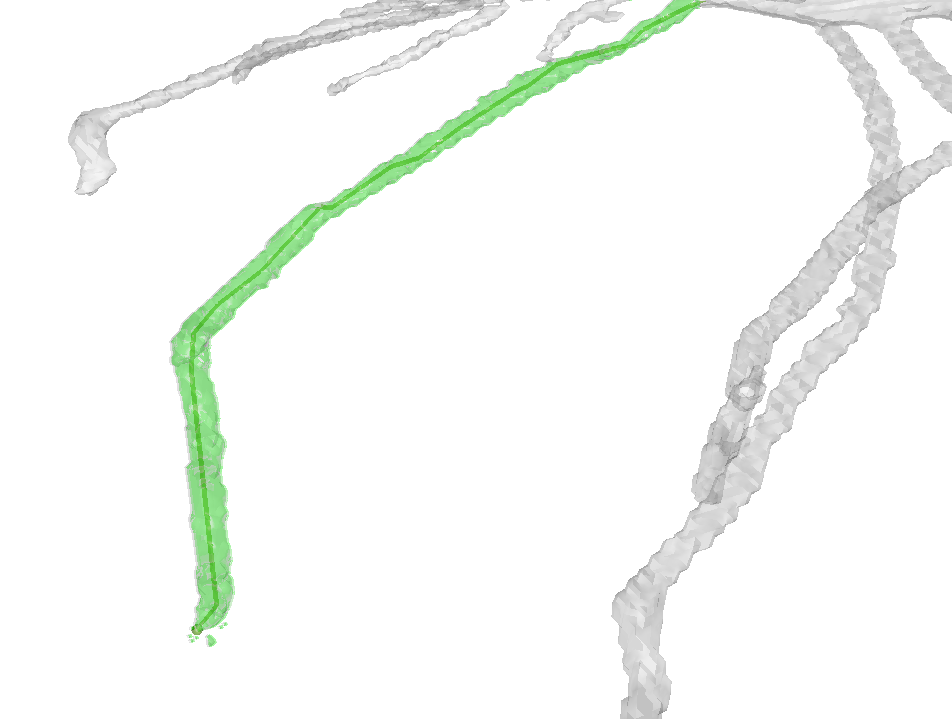}\label{fig:skel_filled}}
		\vspace*{1ex}
		\caption{Example skeletal branch extraction. From a found quench point (a) a path is connected (b) to the start point and the surrounding area is filled (c).}
		\label{fig:skel}	
	\end{figure*}

	Let $\tau(p)=\{p_0, p_1,\ldots, p_{n-1}, p \} \subset I_{seg}$ be the ordered set of voxels constituting the shortest path connecting $p$ to $p_0$. Let $p_n$ be a no-gap position that is explored in the current iteration. Assume $p_{n-1}$ is a gap position, then:
	\begin{align}
		s = \underset{i \in [0,1,\ldots,n-2]}{\text{argmax}} C_{gap}(p_i) \leq \frac{\text{max}(C_{gap})}{2}. \nonumber
	\end{align}
	Obviously, the gap bridged between $p_s$ and $p_n$ has length $n-s$. This length is used for early stopping. If $n-s \leq l_{I{seg}}$ it is assumed that the found gap is missing root information and it is subsequently bridged. The cost of $\tau(p_n)$ is updated to reflect the assumption that the path should have traversed a missing lower-cost area:
	\begin{align}
		C_{\tau}(p_e) = C_{\tau}(p_s) +(n-s)\cdot C_{gap}(p_e). \nonumber
	\end{align}
	
	The updated cost is lower than the initial cost. Hence, $p_e$ is the next position to be expanded. Due to Eq.~\ref{for:gap}, the area around $p_e$ is usually explored before further gaps are found. In practice, this leads to a singular connecting path between disconnected low-cost areas. This avoids the initially described problem of large areas of high cost being included.

	\subsection{LCC Extraction}
	Due to early stopping, $C_{\tau} = \infty$ for all positions for which no path of cost lower $\omega_{I_{seg}}$ can be found. The largest connected component $I_{lcc}$ is now defined as:
	\begin{align}
		ext(p) &= C_{\tau}(p) < \infty \land I_{seg}(p) \geq \gamma_{I_{seg}}, \nonumber \\
		con(p) &= \exists p_t: ext(p_t) \land p \in \tau(p_t),\nonumber \\
		I_{lcc}(p) &= 
		\begin{cases}
			1 & \text{, if } ext(p) \lor con(p)\\ 
			0 & \text{, otherwise} 
		\end{cases}. \nonumber
	\end{align}
	
	The resulting LCC is a binary volume excluding noise clusters far away and disconnected from the starting point, while connecting close-by root parts using a unique connection. This can be seen in Fig.~\ref{fig:run_lcc}.

\section{Curve Skeletonization}

\label{sec:curve}
By construction, $I_{lcc}$ is a single connected structure allowing for extraction of a skeleton, for which we use a modified version of \cite{jin2016robust}. A dilation multiplier $\beta_{I_{seg}} \in [1.1,2.0]$ is given. If interpolation is used, $\delta_{I_{seg}} \in \mathbb{R}$ is given.
	
	\subsection{Centerline Cost Maps}
	The extracted skeleton should follow the centerline of a root branch. To this end, $I_{lcc}$ is used to generate a new radius-based cost map, similar to Eq.~\ref{for:rad_est}:
	\begin{align*}
		\allowdisplaybreaks
		R_{lcc}(p) &= \underset{r \in \mathbb{N}}{\text{argmax}}\left(\frac{|\xi_{I_{lcc},r'}^+(p)|}{|\phi_{r'}|} \geq 0.9, \forall r' \leq r\right), \nonumber\\
		C_{skel}(p,f) &=
			\begin{cases}
				f(p) & \text{, if } R_{lcc} > 0 \\
				\infty & \text{, otherwise}
			\end{cases}, \label{for:skel} \\
		c_{R_{lcc}}(p) &= 1 - \frac{R_{lcc}(p)}{\text{max}(R_{lcc})}, \nonumber \\
		C_{R_{lcc}}(p) &=C_{skel}(p,c_{R_{lcc}}). \nonumber
	\end{align*}
	
	For centerline extraction, only the radius relative to the radii of surrounding positions is of interest. Let $\chi(p)$ be the 26-neighborhood of $p$. Then 
	\begin{align}
		e_{R_{lcc}}(p,p_t) &= 
			\begin{cases}
				1 & \text{, if } R_{lcc}(p) > R_{lcc}(p_t)\\
				0 & \text{, otherwise}
			\end{cases},\nonumber \\
		c_{rel}(p) &= 1- \frac{\sum_{p_t \in \chi(p)} e_{R_{lcc}}(p,p_t)}{26}, \\
		C_{rel}(p) &= C_{skel}(p,c_{rel}). \nonumber
	\end{align}
	Fig.~\ref{fig:run_scost} shows such a cost map.

	\subsection{Quench Points and Shortest Path}
	Using $R_{lcc}$, the list of all quench points $\theta$ is generated:
	\begin{align}
		\theta = \left\{p_q \middle| \underset{p_t \in \chi(p_q)}{\sum}e_{R_{lcc}}(p_q, p_t) > 20 \right\}. 
	\end{align}
	Let $\theta_{sort}=\{q_0, q_1, \ldots\}$ be $\theta$ sorted in decreasing order of Euclidean distance to $p_0$. 
	
	Again, Dijkstra shortest path is used to compute the shortest path $\tau(p)$ for each $p$. By definition of Eq.~\ref{for:skel}, only voxels included in the LCC have valid path costs. Some scans have a large flat area near the plant shoot. A simple $cut_z \in [0,z]$ can be given to ignore all quench points above a height $cut_z$. This excepts the flat non-root area from extraction.

	\subsection{Branch Extraction}
	
	A root graph $G_{full}$ is extracted on a voxel basis. Each voxel in a given branch is included in form of a graph node. A graph node is defined as triple $node(p)=(p,r,id)$ where $p \in \mathbb{R}^3$ is the voxel position, $r = R_{lcc}(p)$ the local radius estimate and $id \in \mathbb{N}$ a branch id unique for each branch. 
	
	Initialize $V_{node}$ and $V_{occ}$ as 0 volumes and $G_{full}$ as root graph with root node $node(p_0)=(p_0, R_{lcc}(p_0),0)$. Set $V_{occ}(p_0) = 1$ and $V_{node}(p_0)= {^*node(p_0)}$ where * denotes the reference operator. These volumes are used to keep track of areas already part of $G_{full}$. Fig.~\ref{fig:skel} shows the process of creating and adding a branch. %Alg.~\ref{alg:skel} shows the skeletonization algorithm.
	
	%\subfile{../algorithms/build_skeleton2}
	
		\subsubsection{Creating a Branch}
		Let $q$ be the quench points furthest from $p_0$ that is not yet part of $G_{full}$ and is not included in an already existing branch. This is the first position $q \in \theta_{sort}$ such that $V_{occ}(q) < 1$. 
		
		Create the node $(q,R_{lcc}(q),id_{run})$, where $id_{run}$ is a running integer that is incremented every time after a new branch is created and extracted. Now, $\tau(q)$ is followed starting from endpoint $q$ to create a graph branch. While $V_{node}(p_i)=0$, a new node $(p_i,R_{lcc}(p_i),id_{run})$ is created and the node corresponding to $p_{i+1}$ is added as a child node. Once $V_{node}(p_j)\neq 0$, the $node(p_{j+1})$ is added as child to the node referenced in $V_{node}(p_j)$.
		
		This results in a connected root branch that grows until it finds an area already part of the root graph. The branch follows the centerline of highest radius, due to using the inverted radius as cost (see Fig.~\ref{fig:skel_extr}).

		\subsubsection{Filling control Volumes}
		\label{sec:graph_fill}
		Whenever a new node at position $p_i$ is created, the surrounding area in $V_{occ}(p_i)$ and $V_{node}(p_i)$ is filled using sphere masks:
		\begin{align}
			r_{dil} &= R_{lcc}(p_i) \cdot \beta_{I_{seg}}, \nonumber\\
			V_{occ}(p) &= 1 &\forall &p \in \xi_{V_{occ},r_{dil}}(p_i), \\
			V_{node}(p) &= {^*node(p_i)} &\forall &p \in \xi_{V_{node},r_{dil}/2}(p_i). \nonumber
		\end{align}
	
		By increasing the local radius estimates, noisy surface areas are included while filling $V_{occ}$ and, hence, quench points found in these areas will not be considered for further branch extraction. This decreases the number of noisy branches. Fig.~\ref{fig:skel_filled}
		shows the resulting filled area in $V_{occ}$ limited to voxels included in the LCC.		
		
	\subsection{Interpolation}
	As voxel-perfect extraction is not necessary, the number of graph nodes can be reduced. To this end, the Douglas-Peucker algorithm \cite{douglas1973algorithms} is employed. All branch endpoints and nodes with more than one child in $G_{full}$ are fixed. The subgraphs between these fixed points are pruned by applying the Douglas-Peucker algorithm with parameter $\delta_{I_{seg}}$. The result is the reduced root graph $G$.

\section{Results}
\label{sec:results}

% here, we can show some results:
% 1. qualitative / quantitative results on synthetic data
% 2. qualitative results on root MRI data

\subsection{Evaluation Method}

%TODO present usual parameterization
\subsubsection{Evaluation Dataset}
To evaluate the performance of the extraction algorithm, 22 real plant root MRI scans are used. For each of these, a manually expert-annotated root graph is provided as target. 

To decrease the noise in the original scan, 3D U-Net based root vs. soil segmentation~\cite{zhao20203d} is applied. A total of five different models are used, with root weights 1, 10, 100, 1000 as well as the Log1 loss modification employed during training. The resulting dataset consists of $22 \cdot 5$ MRI segmentations.

\begin{figure}
	\centering
	\begin{subfigure}[t]{0.20\textwidth}
		\centering
		{\includegraphics[width=\textwidth]{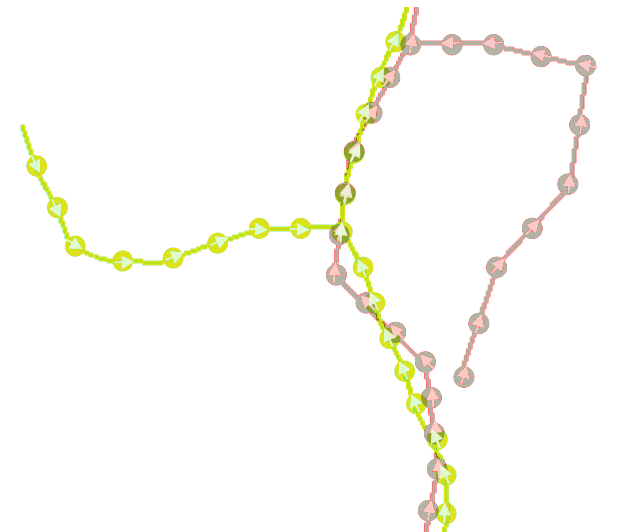}}
		\caption{Extracted structure and manual reconstruction}
	\end{subfigure}\hspace*{2ex}
	\begin{subfigure}[t]{0.20\textwidth}
		\centering
		{\includegraphics[width=\textwidth]{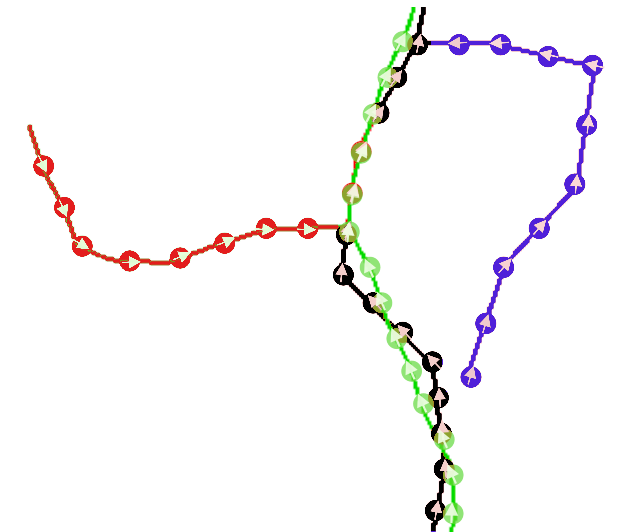}}
		\caption{Extracted structure and manual reconstruction after evaluation is applied}
	\end{subfigure}
	\caption{Yellow: Extraction; Gray: Manual reconstruction; Green/Red: Extraction with/without corresponding manual reconstruction; Black/Blue: Manual reconstruction with/without corresponding extraction.}
	\label{fig:gt_vs_extraction}
\end{figure}

%\begin{table}[h]
%	\centering
%	\caption{Average performance using extraction dependent on segmentation variant. Cost $C_{lcc}$ is used.}
%	\small
%	\begin{tabular}{l||c|c|c}
%		\hline
%		\bfseries Segmentation & \bfseries F1-Score & \bfseries Precision & \bf Recall\\
%		\hline\hline
%		Log1 & 0.8279 & 0.8339 & 0.8312 \\
%		Root W 1 & 0.7765& 0.8911& 0.7005 \\
%		Root W 10 & 0.8194& 0.8465& 0.8019 \\
%		Root W 100 & 0.7683 & 0.7030 & 0.8682 \\
%		Root W 1000 & 0.6973 & 0.5953 & 0.8898 \\
%		\hline
%	\end{tabular}
%	\label{table:models}
%\end{table}

%\begin{table}[h]
%	\centering
%	\caption{Average performance using extraction dependent on segmentation variant. Cost $C_{rel}$ is used.}
%	\small
%	\begin{tabular}{l||c|c|c}
%		\hline
%		\bfseries Segmentation & \bfseries F1-Score & \bfseries Precision & \bf Recall\\
%		\hline\hline
%		Log1 & 0.8294 & 0.8303 & 0.8369 \\
%		Root W 1 & 0.7767& 0.8783& 0.7088 \\
%		Root W 10 & 0.8214& 0.8248& 0.8081 \\
%		Root W 100 & 0.7744 & 0.7040 & 0.8799 \\
%		Root W 1000 & 0.7001 & 0.5958 & 0.8955 \\
%		\hline
%	\end{tabular}
%	\label{table:models2}
%\end{table}

\subsubsection{Distance-tolerant F1 Score for Graphs} %Needs to be emphasized more with respect to 
The following method is used to evaluate the extracted root graph $G$ against the manually reconstructed root $G_T$. Spacing $s$ and distance tolerance $d$ are given as parameters.

Let $l_G$ be a list containing tuples $(p,dir(p))$ for all $p \in G$, where $dir(p)$ is the direction from parent node $p'$ to $p$. If $\left\Vert p,p'\right\Vert_2 > s$, the line segment connecting $p$ and $p'$ is segmented into points $p_1,p_2,\ldots,p_{n-1}$ such that $\left\Vert p_i,p_{i+1}\right\Vert_2 = s$ for all $i \in 0,1,\ldots,n-2$ using $p_0 = p'$. Tuple $(p_i, dir(p))$ for all $i \in 1,\ldots,n-1$ are also added to $l_G$.

This step is repeated for the target graph $G_T$ to generate $l_{T}$. Fig.~\ref{fig:gt_vs_extraction} shows an example of this. As vertex density between extracted graphs varies depending on interpolation or human annotation this step is used to create a dense graph representation.

Two points $p_{T} \in l_{T}$ and $p_{G} \in l_{G}$ correspond if two conditions are met. First, the angle between $dir(p_T)$ and $dir(p_G)$ has to be smaller 90 degrees. The second condition is that $\left\Vert p_T,p_G\right\Vert_2 \leq d$, or $\left\Vert p_T,line(p_G)\right\Vert_2 \leq d$, where $line(p_G)$ is the line connecting $p_G$ and its previous point. 

If correspondence between given $p_{T}$ and multiple $p_{G}$ can be established, the closest point is chosen. Each $p_{G}$ can correspond to at most one $p_{T}$ and corresponding points in $l_G$ are marked as such.

Finally, each point in $l_{G}$ with a corresponding point in $l_{T}$ is regarded as a true positive, each point in $l_{G}$ without a corresponding point in $l_{T}$ is regarded as a false positive, and each point in $l_{T}$ without a corresponding point in $l_{G}$ is regarded as a false negative. This is then used to compute recall, precision, and F1 score.

This method evaluates the extraction with respect to topology by penalizing large divergence in orientation and missing target in the immediate area. Furthermore, the extraction takes branch length into account by penalizing branches which are either too short or too long with respect to the target. As this evaluation scheme does not allow multi-correspondence and the manual reconstruction is not perfectly aligned $d=15$ is chosen for all evaluations.

\subsection{Performance}

\begin{figure*}
	\centering
	\begin{subfigure}[h]{0.22\textwidth}
		\centering
		\includegraphics[width=.95\textwidth]{"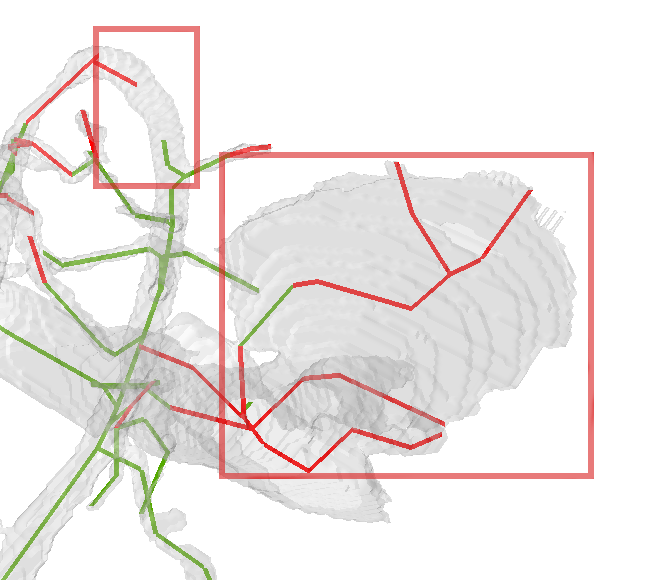"}
		\caption{Left mark: Root cut in half; Right mark: Large flat connected area containing false extraction.}
		\label{fig:rw10_3sdap8}
	\end{subfigure}\hspace*{2ex}
	\hspace{1ex}
	\begin{subfigure}[h]{0.22\textwidth}
		\centering
		\includegraphics[width=.95\textwidth]{"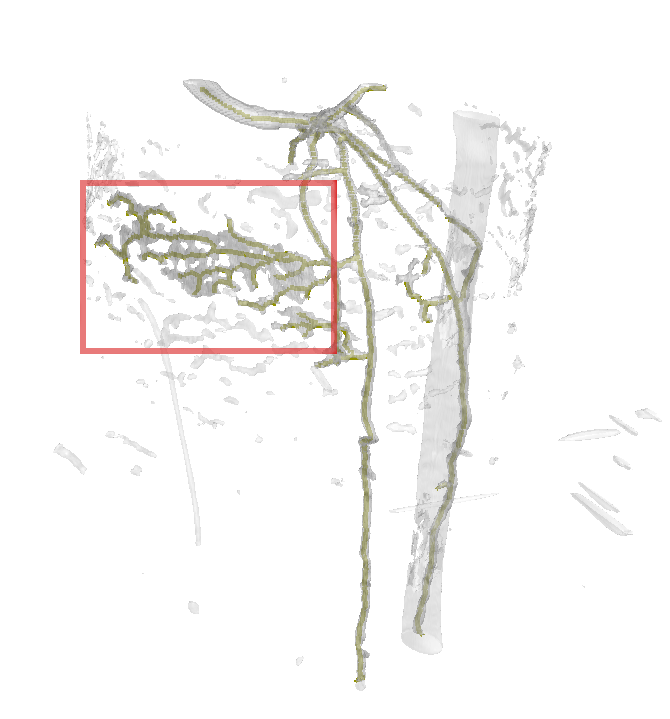"}
		\caption{Mark: Noise close to the root is extracted.}
		\label{fig:noise_in_extr}
	\end{subfigure}
	\hspace{1ex}
	\begin{subfigure}[h]{0.4\textwidth}
		\centering
		\includegraphics[width=.95\textwidth]{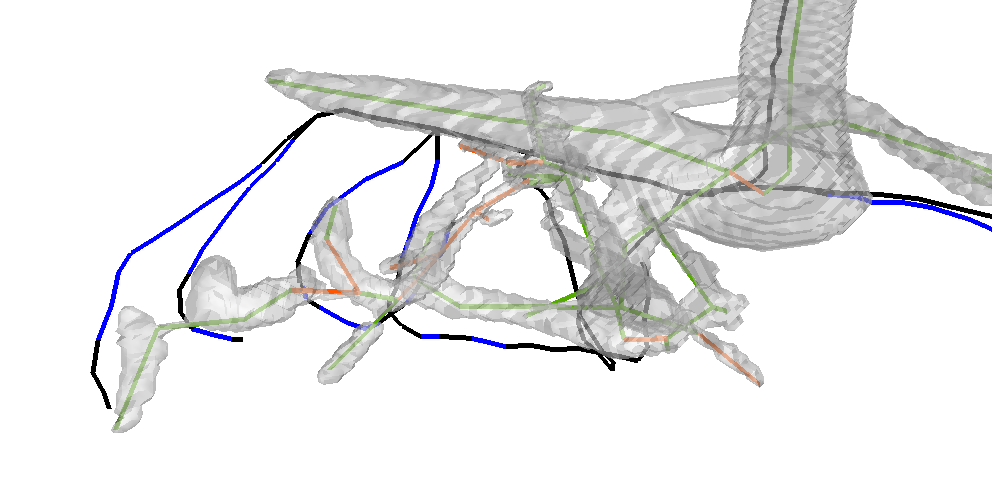}\vspace*{-2ex}
		\caption{The manual reconstruction shows intricate arc-like structures not found in the segmentation. The missing volume information lead to wrong extraction.}
		\label{fig:3sand1w8}
	\end{subfigure}
	\vspace*{1ex}
	\caption{Green/Red: Extraction with/without corresponding manual reconstruction; Black/Blue: Manual reconstruction with/without corresponding extraction.}
\end{figure*}

\begin{table}[!t]
	\centering
	\caption{Average performance using extraction dependent on segmentation variant and cost maps.}
	\small
	\begin{tabular}{l|l||c|c|c}
		\hline
		\bfseries Cost & \bfseries Segmentation & \bfseries F1-Score & \bfseries Precision & \bf Recall\\
		\hline\hline 
		\multirow{5}{*}{$C_{R_{lcc}}$}
		&Log1 & 0.8279 & 0.8339 & 0.8312 \\[0.1ex]
		&Root W 1 & 0.7765& 0.8911& 0.7005 \\
		&Root W 10 & 0.8194& 0.8465& 0.8019 \\
		&Root W 100 & 0.7683 & 0.7030 & 0.8682 \\
		&Root W 1000 & 0.6973 & 0.5953 & 0.8898 \\
		\hline\hline 
		\multirow{5}{*}{$C_{rel}$}
		&Log1 & 0.8294 & 0.8303 & 0.8369 \\
		&Root W 1 & 0.7767& 0.8783& 0.7088 \\
		&Root W 10 & 0.8214& 0.8248& 0.8081 \\
		&Root W 100 & 0.7744 & 0.7040 & 0.8799 \\
		&Root W 1000 & 0.7001 & 0.5958 & 0.8955 \\
		\hline
	\end{tabular}
	\label{table:models}
\end{table}

\begin{figure}[!t]
	\centering
	\begin{subfigure}[b]{0.3\textwidth}
		\centering
		\includegraphics[width=\textwidth]{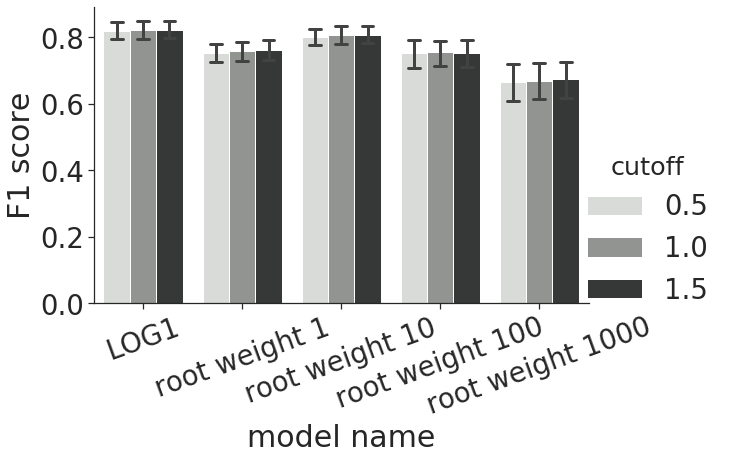}
		\caption{Changing $\omega_{I_{seg}}$}
	\end{subfigure} 	\vspace*{.1ex}
	%\hspace{2ex}
	\begin{subfigure}[b]{0.3\textwidth}
		\centering
		\includegraphics[width=\textwidth]{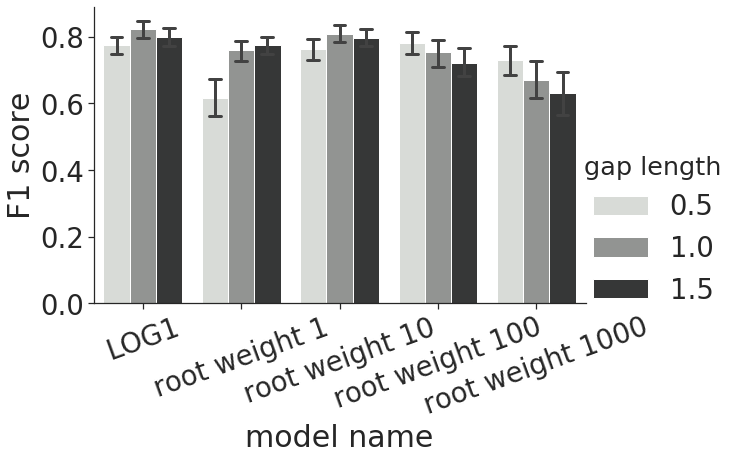}
		\caption{Changing $l_{I_{seg}}$} %\vspace*{1ex}}
		\label{fig:params_gl}
	\end{subfigure}	\vspace*{.1ex}
	%\hspace{2ex}
	\begin{subfigure}[b]{0.3\textwidth}
		\centering
		\includegraphics[width=\textwidth]{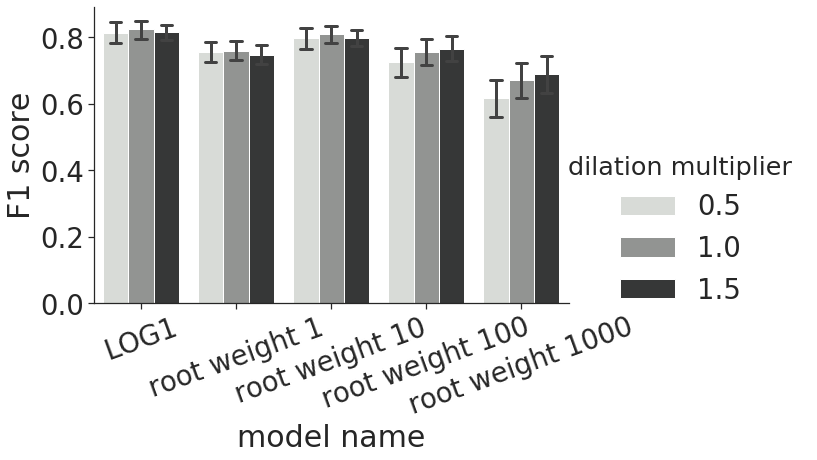}
		\caption{Changing $\beta_{I_{seg}}$}
		\label{fig:params_dl}
	\end{subfigure}
	%\vspace*{1ex}
	\caption{Average performance of the extraction over the complete dataset when changing one of the parameters. No $cut_z$ is used for these values.}
	\label{fig:params_perfomance}
\end{figure}

%\begin{figure*}[!t]
%    \centering
%    \subfloat  
%        [Each scan placed on the recall and precision grid with respect to the model used for segmentation. Precision and Recall are taken from highest F1 score extraction for each scan and model respectively.]
%        {\includegraphics[width=.35\textwidth]{imgs/rec_pre_individual_data_scatterplot.png}\label{fig:scatter_inps}}
%    \hspace{.05\textwidth}
%    \subfloat
%        [Average performance per model using different parameter permutations.]
%        {\includegraphics[width=.35\textwidth]{imgs/rec_pre_diff_params_lineplot.png}\label{fig:overview_inps}}
%    \caption{Comparison of recall and precision for different segmentation models}
%\end{figure*}

\begin{figure*}[!t]
	\centering
	\subfloat[Full volume, extraction and manual reconstruction]{\includegraphics[width=.35\textwidth]{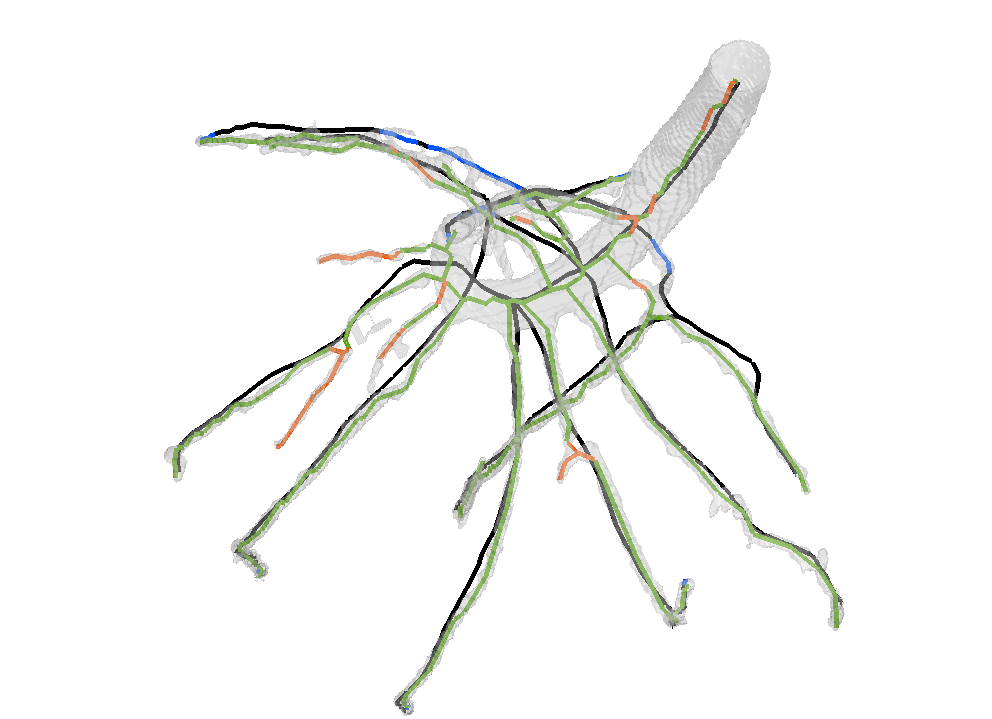}\label{fig:4soil3d6_full}}
	\hfill
	\subfloat[The marked areas contain elongated volumes resembling roots. These are not part of the target structure but may have been missed during manual extraction]{\includegraphics[width=.30\textwidth]{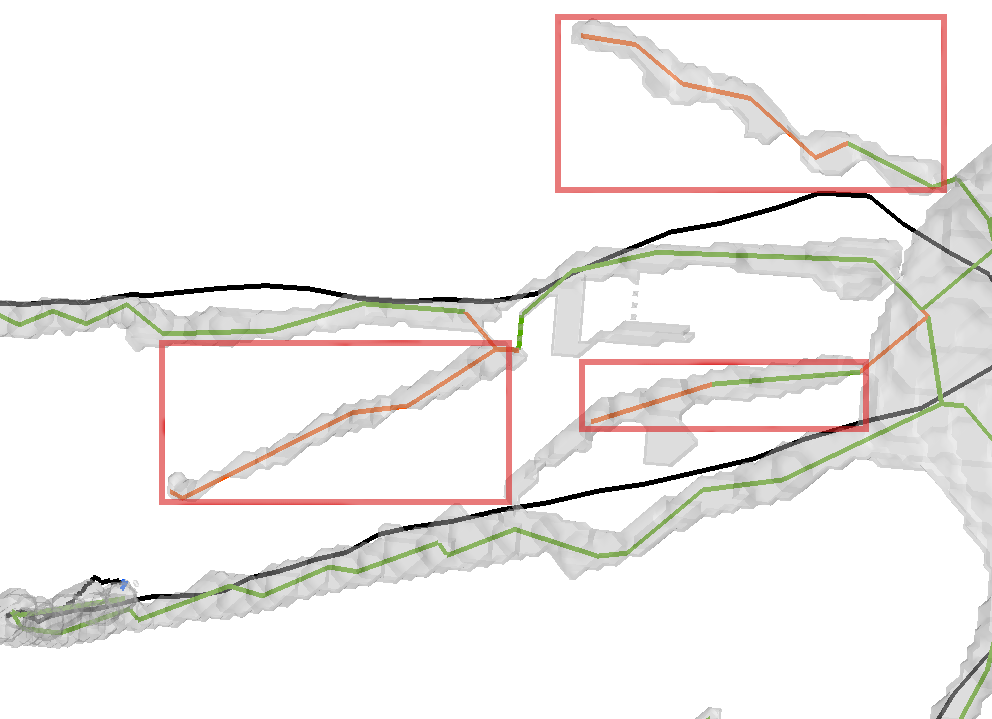}\label{fig:4soil3d6_no_gt}}
	\hfill
	\subfloat[Top: Manual reconstruction; Bottom: Extracted Structure; Marked area: Lower branch is split and wrongly connected to the upper branch.]{\includegraphics[width=.30\textwidth]{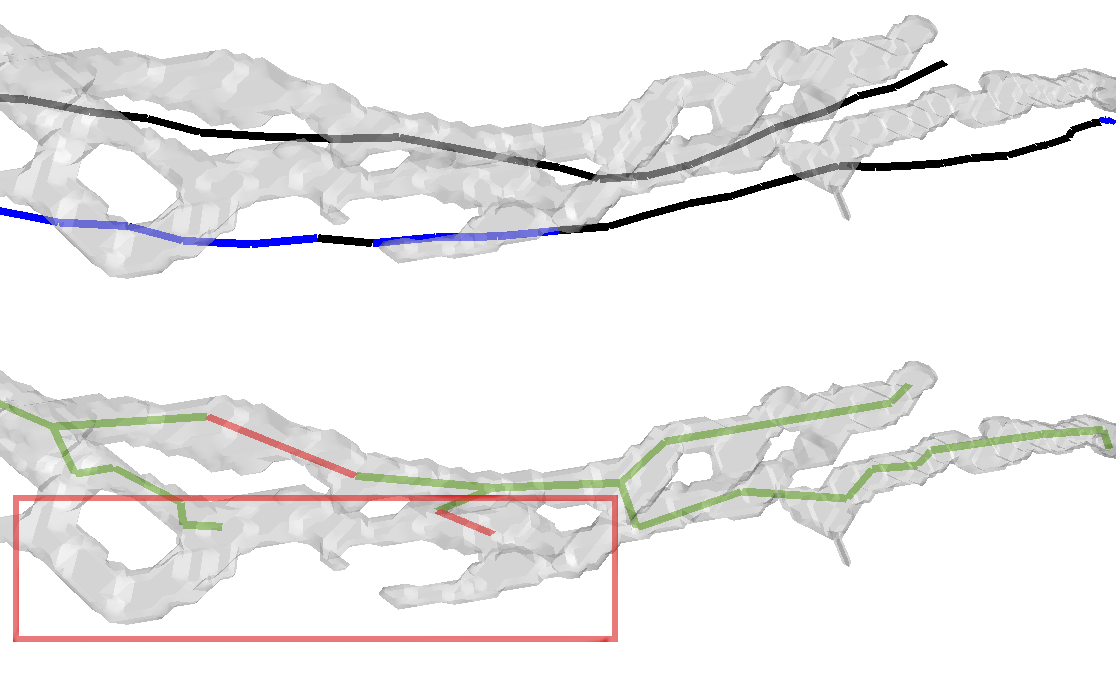}\label{fig:4soil3d6_merge}}
	\vspace*{1ex}
	\caption{Extracted root graph based on Fig.~\ref{fig:run_cskel} and corresponding manual reconstruction. Green/Red: Extraction with/without corresponding manual reconstruction; Black/Blue: Manual reconstruction with/without corresponding extraction.}
	\label{fig:4soil3d6}	
\end{figure*}

\subsubsection{Performance based on Input Quality}

The performance of our algorithm strongly depends on input quality. Consequently, the quality of the initial U-Net based segmentation heavily determines the performance of the root extraction algorithm. Table~\ref{table:models} shows the average F1 score of the extraction algorithm dependent on the segmentation variant. As can be seen, the Log1 loss modification gives the best F1 score by producing a balanced precision and recall.

The second-best model utilizes a root-loss weight of 10. The segmentation outputs have slightly enlarged root structures and noise structures. For most roots, the score is slightly lower than the Log1 score, but the enlarged areas result in merged roots leading to wrong connection in more dense root systems. Another issue is an enlarged area towards the plant shoot of a root system leading to multiple wrong extractions reducing the precision noticeably. Fig.~\ref{fig:rw10_3sdap8} shows this issue. This is addressed by employing $cut_z$.

Using a root weight of 1 combined with the same parametrization used for the other models results in noticeably lower recall. Further parameter testing showed that a much larger maximum gap length is needed here. Using this, the recall can be increased but still stays below the other models.

The segmentation models utilizing larger root weights 100 and 1000, respectively, show problems similar to root weight 10. More dense root systems tend to merge resulting in structures combining two root branches that should be separate. Another issue is that the enlarged roots and noise areas merge in some scans leading to noise areas being treated as root, see Fig.~\ref{fig:noise_in_extr}. These noise areas are large enough to have a non-trivial radius, limiting the use of radius-based penalties.

Table~\ref{table:models} also shows the average performance over all scans with respect to the employed cost maps $C_{R_{lcc}}$ and $C_{rel}$. As can be seen, $C_{rel}$ increases recall while decreasing precision. Root scans containing thicker roots like segmentation using rw100 or rw1000 show a slight increase in precision as well. This is due to the relative cost not suffering from radius cost saturation for large radii.

\subsubsection{Performance based on Parameterization}

Extraction performance was evaluated also with respect to three parameters that were shown to change extraction behavior the most in manual assessment. Fig.~\ref{fig:params_perfomance} shows the behavior of the extraction when altering these parameters. Starting from a manually found parameter configuration, for each file these parameters are tested using the multipliers 0.5, 1.0, and 1.5.

Altering the cost cutoff $\omega_{I_{seg}}$ changes behavior only minimally once above a certain threshold. Below this threshold, $\omega_{I_{seg}}$ can be used to exclude higher-cost areas from the extraction. This only works if unwanted areas can be sufficiently penalized, which is not the case for models with high root weighting. The threshold can vary from 100 in scans with overall high connectivity and large radius to 1500 in scans containing long and thin root systems. This may lead to the changes in $\omega_{I_{seg}}$ having only a small effect on performance.

Changes in gap length $l_{I_{seg}}$ have a larger effect on performance. Well-chosen $l_{I_{seg}}$ can increase the performance significantly. This parameter depends on the input scan quality. As can be seen in Fig.~\ref{fig:params_gl}, $l_{I_{seg}}$ used and validated on models Log1 and root weight 10 is too large for the models using higher root weighting while the performance with root weight 1 segmentation increases noticeably with larger $l_{I_{seg}}$.

The third parameter evaluated is the dilation multiplier $\beta_{I_{seg}}$ for root extraction. Models with smaller more discernible structures, as can be found in root weights 1, 10 and Log1, show small change when changing $\beta_{I_{seg}}$. These inputs have clearer structures leading to fewer false responses in surrounding areas. Models with high root weights often have areas creating quench points which are wrongfully included because of insufficient radius dilation during skeletonization. This can be reduced by increasing $\beta_{I_{seg}}$ as can be seen in Fig.~\ref{fig:params_dl}. Hence, this parameter can be used to increase precision in more noisy scans.

\subsection{Qualitative Assessment}
%\begin{figure}
%    \centering
%    \begin{subfigure}[b]{.4\textwidth}
%        \includegraphics[width=\textwidth]{"imgs/fig6/close_up.png"}
%    \end{subfigure}
%    \caption{. Blue Area: Missing extraction, instead of a single root, some quench points are connected to the upper branch.}
%    \label{fig:4soil3d6}
%\end{figure}

%\begin{figure}
%	\centering
%	\begin{subfigure}[b]{.4\textwidth}
%		\includegraphics[width=\textwidth]{"imgs/fig6/close_no_gt_b.png"}
%	\end{subfigure}
%	\caption{Black: Ground truth segments with corresponding extraction output. Green: extraction with corresponding ground truth, Red: extraction without corresponding ground truth. The marked areas contain elongated volumes resembling roots. These are not part of the target structure but may have been missed during manual extraction.}
%	\label{fig:4soil3d6_no_gt}
%\end{figure}

%\begin{figure}
%	\centering
%	\includegraphics[width=.40\textwidth]{imgs/fig7/missing_str.png}\vspace*{-2ex}
%	\caption{Green/Red: Extraction with/without corresponding manual reconstruction; Black/Blue: Manual reconstruction with/without corresponding extraction. The manual reconstruction shows intricate arc-like structures not found in the segmentation. The missing volume information lead to wrong extraction.}
%	\label{fig:3sand1w8}
%\end{figure}

Fig.~\ref{fig:4soil3d6_full} shows a scan with high F1 score. The overall structure is extracted correctly. Some areas that appear to have an elongated although fractured structure are extracted as root while not being part of the target structure, see Fig.~\ref{fig:4soil3d6_no_gt}. These may be roots that did not get annotated during manual extraction. While fractured, the extraction creates a single connected response---bridging gaps correctly.

Some structures in the segmentation merge. This leads the extraction algorithm to connect merged areas using a single branch instead of the two or more that should be used. Fig.~\ref{fig:4soil3d6_merge} shows this. Two parallel branches merge in certain intervals leading to the lower branch not being extracted as a single branch but as segments connected to the upper branch.

This can also be observed in Fig.~\ref{fig:rw10_3sdap8}. Towards the upper part, a root is clearly cut in half during extraction. This is due to the same structure merging creating wrong paths.

In Fig.~\ref{fig:3sand1w8}, the area towards the center left has missing structure. The target shows curved downwards connections not found in the segmented MRI. Instead, the lower parts are connected by the shortest gap.

As can be seen in Fig.~\ref{fig:intr_extr}, our algorithm is capable of extracting root structures from large complicated root scans. Disconnected roots have been connected correctly.

\subsection{Runtime and Memory Usage}
\label{sec:rt}
\begin{figure}
    \centering
    \includegraphics[width=.43\textwidth]{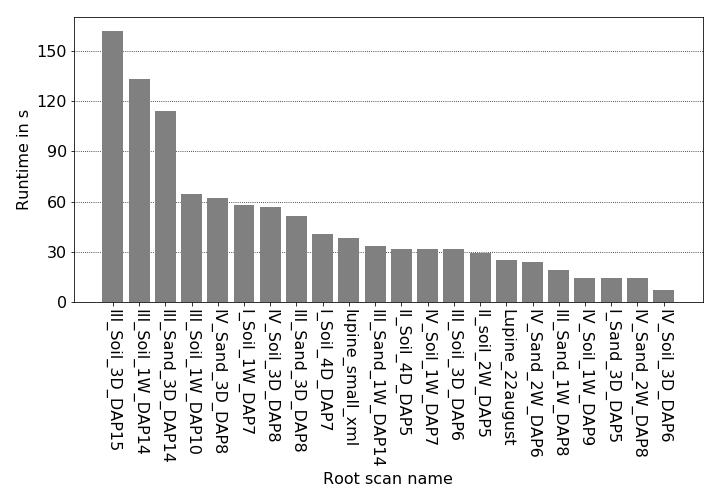}
    \caption{Average runtime of the full algorithm dependent on input scan.}
    \label{fig:runtime}
\end{figure}

The test computer is a laptop with two cores at 2.3\,GHz and 8\,GB of RAM running Ubuntu 16.04 64-bit. Input size is between 140$\times$512$\times$512 to 396$\times$512$\times$512. Maximum memory usage is 10.9\,GB. Average memory consumption is 4-6\,GB. The three scans needing the most time in Fig.~\ref{fig:runtime} need more than the available RAM. The increase in runtime is because of paging out of memory used by the algorithm.

The other relevant factors determining the runtime are the maximum radius in structures and the amount of extracted roots.

\section{Discussion and Future Work}

% ... we presented a novel algorithm...
% benefits...
% TODO more information of what has been presented
In this paper, a pipeline to extract root structures from MRI images containing noise and disconnected structures has been presented. The pipeline takes a two-stage approach to first reduce noise and connect disconnected areas, followed by curve skeletonization to extract a root skeleton from the scan. 

The developed method runs on a laptop with limited resources and is capable of processing large datasets of input MRI scans quick enough for interactive parametrization and visualization.

Evaluation on real MRI scans shows that performance depends on the quality of the input. Segmentations using Log1 loss modification and a slightly higher root weighting perform the best. These segmentations preserve most of the root structure while not enlarging them too much.

Two input properties negatively affecting extraction have been observed: The first one is large regions of missing information in the input. This may lead to wrong connections, as the algorithm takes the shortest path through the remaining volume.

The second property concerns merging root structures. The current algorithm does not consider local branch orientation, which could potentially be used to address the issue of merging roots as well as help with guessing correct connections missing in the input.

Overall, the algorithm is capable of extracting complex root structure from scans with low connectivity and high noise. It produced some root branches not found in the manual reconstruction, demonstrating its potential for improving not only speed, but also the quality of the reconstruction. 

In future work, an iterative approach to graph extraction could be investigated. Instead of reconstructing the full root structure graph in a single skeletonization run, the method could be initialized by reconstructing only the most obvious roots first. The reconstruction could then be iteratively extended by further skeletonization runs. This would allow for using initial reconstructions as context for the interpretation of the more ambiguous MRI regions. Such an approach could also be used to improve finding and connecting some disconnected branches where the shortest path method for filling gaps is less suitable.

\subsection*{Acknowledgment}
This research was supported by grants BE 2556/15 and SCHN 1361/3-1 of German Research Foundation (DFG).

%\newpage
\bibliographystyle{IEEEtran}
\bibliography{IEEEabrv, bibliography}

\end{document}